\documentclass[journal]{IEEEtran}
\usepackage{amsmath,amsfonts}
\usepackage{algorithm}
\usepackage{algorithmicx}
\usepackage{algpseudocode}
\usepackage{verbatim}
\usepackage{graphicx} 
\usepackage{url}
\usepackage{booktabs, subfig}
\usepackage[table]{xcolor}
\usepackage{threeparttable}
\usepackage{booktabs}
\usepackage{tabularx}
\usepackage{arydshln}
\usepackage{multirow}
\usepackage{subfig}
\usepackage{threeparttable}
\usepackage[pagebackref=false,breaklinks=true,colorlinks,bookmarks=false,urlcolor=magenta]{hyperref}
\usepackage[numbers,sort&compress]{natbib}
\definecolor{lightgreen}{RGB}{169,209,142}
\definecolor{lightblue}{RGB}{155,192,226}
\definecolor{lightorange}{RGB}{244,177,131}
\definecolor{lightyellow}{RGB}{255,217,102}

\begin{document}

\title{XTQA: Span-Level Explanations for Textbook Question Answering}

\author{
	Jie Ma, \IEEEmembership{Member,~IEEE,}
	Qi Chai,
	Jun Liu, \IEEEmembership{Senior Member,~IEEE,}
	Qingyu Yin,
	Pinghui Wang, \IEEEmembership{Senior Member,~IEEE,}
	and Qinghua Zheng, \IEEEmembership{Member,~IEEE,}
	\thanks{Jie Ma and Pinghui Wang are with the Ministry of Education of Key Lab for Intelligent Networks and Network Security, School of Cyber Science and Engineering, Xi'an Jiaotong University, Xi'an, Shaanxi 710049, China (email: jiema@xjtu.edu.cn).}
	\thanks{Jun Liu, Qi Chai, and Qinghua Zheng are with Shaanxi Provincial Key Laboratory of Big Data Knowledge Engineering, the School of Computer Science and Technology, Xi'an Jiaotong University, Xi'an, Shaanxi 710049, China.}
	\thanks{Qingyu Yin is an applied scientist at Amazon.}
}

\markboth{Journal of \LaTeX\ Class Files,~Vol.~14, No.~8, August~2021}%
{Shell \MakeLowercase{\textit{et al.}}: A Sample Article Using IEEEtran.cls for IEEE Journals}



\maketitle

\begin{abstract}
Textbook Question Answering (TQA) is the task of correctly answering diagram or non-diagram questions given large multi-modal contexts consisting of abundant essays and diagrams. In real-world scenarios, an explainable TQA system plays a key role in deepening humans' understanding of learned knowledge. However, there is no work to investigate how to provide explanations currently. To address this issue, we devise a novel architecture towards span-level eXplanations for TQA (XTQA). In this paper, spans are the combinations of sentences within a paragraph. The key idea is to consider the entire textual context of a lesson as candidate evidence, and then use our proposed coarse-to-fine grained Explanation Extracting (EE) algorithm to narrow down the evidence scope and extract the span-level explanations with varying lengths for answering different questions. The EE algorithm can also be integrated into other TQA methods to make them explainable and improve the TQA performance. Experimental results show that XTQA obtains the best overall explanation result (mIoU) of $52.38\%$ on the first $300$ questions of CK12-QA test splits, demonstrating the explainability of our method (non-diagram: $150$ and diagram: $150$). The results also show that XTQA achieves the best TQA performance of $36.46\%$ and $36.95\%$ on the aforementioned splits respectively. We have released our code in \url{https://github.com/dr-majie/opentqa}.
\end{abstract}


\section{Introduction}
\label{introduction}
\begin{figure}[tbp]
	\centering
	\includegraphics[width=\linewidth]{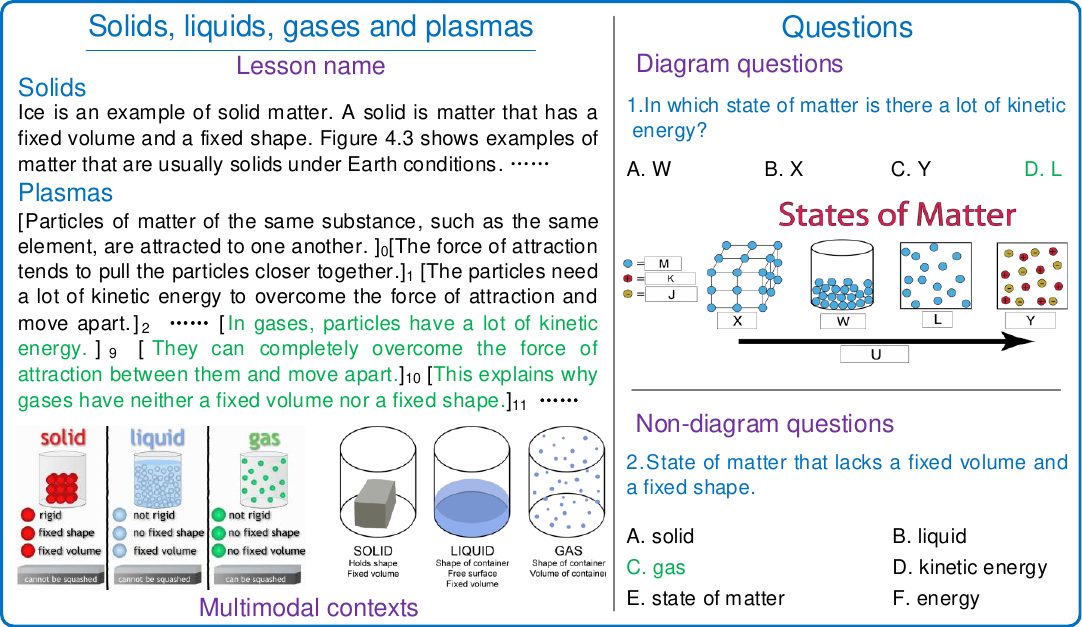}
	\caption{An example of the TQA task. Questions with or without diagrams are shown on the right. Sentences marked in green on the left are the explanations for answering question \emph{1}. The number indicates the order of sentences within a passage.}
	\label{fig.1}
\end{figure}

\IEEEPARstart{Q}{uestion} answering tasks such as Visual Question Answering (VQA) \citep{yu2019deep, khademi2020multimodal} and Machine Reading Comprehension (MRC) \citep{nie2019revealing, saxena2020improving} have attracted the extensive interest of researchers, due to their numerous real-world applications such as intelligent assistants. Recently, a new task called Textbook Question Answering (TQA) \citep{kembhavi2017you} was proposed and it requires a system to answer diagram and non-diagram questions automatically given large multi-modal contexts consisting of abundant essays and diagrams. Different from VQA and MRC, TQA uses both text and diagram inputs in the context and the question, which makes it a non-trivial task. Figure \ref{fig.1} shows an example of the TQA task. In this example, a TQA system is required to provide the answers to questions for humans after learning the multi-modal context of lesson ``Solids, liquids, gases and plasmas" on the left. Humans will be perplexed in real-world education if they are only given the answers because they may not fully comprehend the knowledge involved in the questions. As a result, a desirable TQA system should provide answers as well as explanations for humans, allowing them to gain a better understanding of their learned knowledge. Although existing works \cite{li2018textbook, kim2019textbook, gomez2020isaaq} have made significant progress on the TQA performance, there is currently no work to investigate how to provide explanations to the best of our knowledge.
A recent study \citep{kembhavi2017you} found that about $80\%$ of the questions can be answered by using sentences in the context and we notice that these evidence spans (combinations of sentences in a paragraph) can also be regarded as explanations because they contain the key knowledge to answer the questions. For example, the span $[9, 11]$ marked in green on the left of Figure \ref{fig.1} can be provided for humans to explain why the TQA system chooses \emph{D} for question \emph{1}, where 9 and 11 denotes the start and end indices respectively. 
	
Inspired by this, we devise a novel architecture towards span-level eXplanations for Textbook Question Answering (XTQA), which considers the entire textual context of a lesson as candidate evidence and extracts span-level explanations using our proposed coarse-to-fine grained Explanation Extracting (EE) algorithm. Concretely, we regard each paragraph of a lesson as a document and apply the embedding based query expansion method \cite{kuzi2016query} to choose the top $M$ paragraphs that are relevant to questions in the coarse-grained phase. In the fine-grained phase, the top $U$ span-level explanations are extracted from all candidate spans within the top $M$ paragraphs by computing the information gain of each span for questions. The larger information gain indicates the more uncertainty of questions reduced by spans. We consider the explanations extracted by EE to be the key context for assisting XTQA in predicting answers and improving performance. Due to the lack of ground truth for explanations and their importance for TQA, we use the answer label to optimize explanation extracting. Furthermore, the EE algorithm can also be integrated into other methods to make them explainable and improve the TQA performance.
	
Experimental results show that XTQA obtains the best overall explanation result of $52.38\%$ on the first $300$ questions of the CK12-QA\footnote{The TQA dataset is collected from \url{http://www.ck12.org}. In this paper, we call the TQA dataset CK12-QA to distinguish TQA tasks from TQA datasets.} test split (non-diagram questions: $150$ and diagram questions: $150$) \cite{kembhavi2017you}. The results also show that XTQA achieves the best TQA performance, improving the accuracy on the test split from $34.06\%$ to $36.95\%$.
	
In summary, our contributions are mainly threefold.
\begin{itemize}
	\item[1)] We devise a novel TQA architecture, which considers the entire textual context of a lesson as candidate evidence and extracts span-level explanations with varying lengths for different questions. To the best of our knowledge, this paper is the first work to investigate the explainability of TQA.
		
	\item[2)] We propose the EE algorithm, which can also be integrated into other TQA methods to make them explainable and improve the TQA performance. 
		
	\item[3)] We conduct extensive experiments to explore how well XTQA and baselines+EE provide explanations and how much performance that they can obtain on CK12-QA \cite{kembhavi2017you}. Extensive ablation studies and discussions are also carried out to analyze XTQA.
\end{itemize}

The remainder of this paper is organized as follows. We introduce the related works in Section \ref{relatedwork}. Section \ref{task-for} describes the task formulation. We introduce the details of XTQA in Section \ref{method}. The experiments are discussed in Section \ref{experiment}. Section \ref{conclusion} introduces the concluding remarks.

\section{Related Work}
\label{relatedwork}
In this section, we introduce the related works of three question answering tasks including TQA, MRC and VQA due to their similarities.

\subsection{TQA}
\label{tqa}
There have a few works \cite{li2018textbook, kim2019textbook, gomez2020isaaq} to study TQA. Li \emph{et al.} \cite{li2018textbook} proposed instructor guidance with memory networks, which find contradictions between options and textual context to predict answers. Kim \emph{et al.} \cite{kim2019textbook} proposed a fusion GCN to extract knowledge features and a self-supervised learning method to solve out-of-domain problems. Both of the above papers did not report the accuracy of the test split and release their codes. Ma \emph{et al.} \cite{ma2021relation} proposed a relation-aware fine-grained reasoning network, which builds diagram graphs based on dependency analyses and then applies question-guided attention mechanisms to reason over the graphs. Ma \emph{et al.} \cite{9794290} proposed a weakly supervised multitask learning framework for TQA to strengthen the diagram and text understanding. G{\'o}mez-P{\'e}rez \emph{et al.} \cite{gomez2020isaaq} proposed a pre-trained TQA method ISAAQ based on transformer language models and top-down attention \cite{anderson2018bottom} to solve the multi-modality understanding issue. They pre-trained the textual ISAAQ on RACE \cite{lai2017race}, ARC-Easy, ARC-Challenge \cite{clark2018think} and OpenBookQA \cite{mihaylov2018can} datasets and fine-tuned it on CK12-QA. Similarly, they pre-trained the multi-modal ISAAQ on VQA abstract scenes, VQA \cite{antol2015vqa} and AI2D \cite{kembhavi2016diagram} datasets and fine-tuned it on CK12-QA. By comparison, XTQA tries to provide explanations for humans and it is trained only on CK12-QA.

\subsection{MRC}
MRC requires a machine to answer questions accurately given a textual context \citep{lehnert1977process}. We classify MRC methods into two categories: single-hop and multi-hop reasoning.

\emph{Single-hop methods} \cite{SeoKFH17, YuanFCTPT20, WZDZ020} use specific means such as attention mechanisms to perform interactions between queries and single paragraphs to predict answers. {SeoKFH17} proposed the bi-directional attention flow network to learn query-aware context representations without the early summarization. Yuan \emph{et al.} \cite{YuanFCTPT20} reframed current static MRC environments as interactive and partially observed environments by restricting the context which a model observes at one time and used reinforcement learning to optimize the information-seeking agent. Zhang \emph{et al.} \cite{WZDZ020} integrated the syntactic dependency of interest design into the self-attention network to strengthen the capacity of modeling the linguistic knowledge. However, the answers are very likely to be obtained from multiple paragraphs in real-life scenarios.

\emph{Multi-hop methods} \cite{ding2019cognitive, nie2019revealing, TangSMXYL20, hu2019multi} performs interactions between queries and multiple paragraphs to predict answers. Ding \emph{et al.} \cite{ding2019cognitive} proposed CogQA that builds a cognitive graph by an implicit extraction module and an explicit reasoning module to address the multi-hop question answering. Nie \emph{et al.} \cite{nie2019revealing} proposed a hierarchical pipeline model that reveals the importance of semantic retrieval to give general guidelines on the system design for MRC. Tang \emph{et al.} \cite{TangSMXYL20} proposed a path-based graph convolutional network to perform multi-path reasoning. Hu \emph{et al.} \cite{hu2019multi} proposed a multi-type multi-span network, which combines a multi-type answer predictor with a multi-span extraction method to enhance the MRC performance. In comparison, XTQA extracts evidence spans not only to enhance the TQA performance but also to provide span-level explanations for humans. The spans in this paper are the combinations of sentences rather than words \cite{hu2019multi}.

\subsection{VQA}
VQA requires a machine to answer questions accurately given an image \citep{antol2015vqa}. We classify VQA methods into three categories: joint embedding-based, attention mechanism-based and explainable.

\emph{Joint embedding-based} \cite{fukui2016multimodal, KimOLKHZ17, yu2017multi} methods use convolutional neural networks and recurrent neural networks to learn representations of images and questions respectively, and then project them into a common space to predict answers. Fukui \emph{et al.} \cite{fukui2016multimodal} proposed the multi-modal compact bi-linear pooling method to learn joint input representations and project them into the answer space to predict answers. However, bi-linear representations may limit the applicability to high-dimensional complex computation tasks. To address this issue, the low-rank bi-linear pooling method using Hadamard product \citep{KimOLKHZ17} and factorized bi-linear pooling \citep{yu2017multi} methods are proposed to learn multi-modal representations efficiently. However, these methods may feed irrelevant or noisy information into the answer space.

\emph{Attention mechanism-based} \cite{khademi2020multimodal, anderson2018bottom, li2019relation} methods assign different importance to input representations before information fusion by computing attention coefficients. Anderson \emph{et al.} \cite{anderson2018bottom} proposed a combined bottom-up and top-down attention mechanism that computes attention at the level of salient image regions and objects. Li \emph{et al.} \cite{li2019relation} proposed a relation-aware graph attention network to learn question-adaptive multi-modal representations. Khademi \emph{et al.} \cite{khademi2020multimodal} devised a multi-modal neural graph memory network to perform reasoning about the interactions of objects. Ma \textit{et al.} \cite{9525040} proposed a multitask learning framework to jointly optimize multi-modal learning. However, these methods cannot provide explanations for humans.

\emph{Explainable} methods \cite{wang2018fvqa, huk2018multimodal, yi2018neural, mao2018neuro} give humans explanations with the help of specified means such as external knowledge and symbols. FVQA \citep{wang2018fvqa} queries the external knowledge base to obtain a supporting fact and predicts the answer. Huk Park \emph{et al.} \cite{huk2018multimodal} proposed a multi-modal approach to explanations using post-hoc justifications. Yi \emph{et al.} \cite{yi2018neural} proposed a neural-symbolic visual question answering architecture that disentangles question and image understanding from reasoning. Based on this paper, Mao \emph{et al.} \cite{mao2018neuro} proposed a neuro-symbolic reasoning module that executes generated programs on the latent scene representations to perform reasoning. The explanations of the above works are generated or extracted by complete supervision. By comparison, our model extracts span-level explanations with different lengths for different questions under the answer supervision rather than span supervision, \emph{i.e.}, weak supervision.

\section{Task Formulation}
\label{task-for}
The TQA task can be classified into two categories: diagram question answering and non-diagram question answering. In this Section, we mainly introduce the task formulation of the diagram question answering due to their similarities.

Given a dataset $\mathcal{S}$ consisting of $n$ quadruples $(c_i, q_i, d_i, \mathcal{A}_i)$ with $c_i \in \mathcal{C}$ representing multi-modal contexts of a lesson, $q_i \in \mathcal{Q}$ representing a question, $d_i \in \mathcal{D}$ representing a diagram of $q_i$ and $\mathcal{A}_i \in \mathcal{A}$ representing candidate answers of $q_i$, the task can be denoted as follows:
\begin{equation}
\hat{a}_{i} = \mathop{\arg\max}_{a_{i,j} \in \mathcal{A}_i} p\big(a_{i,j}|c_i, q_i, d_i; \theta\big),
\end{equation}
where $\hat{a}_{i}$ is the predicted answer, $a_{i,j} \in \mathcal{A}_i$ denotes the $j$-th candidate answer of $q_i$, and $\theta$ denotes the trainable parameters. The dataset usually lacks the annotations for span explanations. Therefore, we apply the answer supervision $a_i$ to optimize the explanation extracting. 

In this paper, we only consider the textual context within $c_i$ due to the lack of visual context in some lessons. $ N = \frac{L(L+1)}{2}$ is the number of candidate evidence spans supposing $c_i$ containing one paragraph with $L$ sentences. Candidate evidence span $e_{i,k}$ of $q_i$ is represented by its start $\mathrm{START}(k)$ and end $\mathrm{END}(k)$ indexes respectively following \citep{ma2020jointly}, where $1 \leq k \leq N$, and $1 \leq \mathrm{START}(k) \leq \mathrm{END}(k) \leq L$. For example, if a paragraph is consisting of three sentences with $1, 2, 3$ denoting their indexes, there have $6$ candidate spans including $\{1\}$, $\{1,2\}$, $\{1,2,3\}$, $\{2\}$, $\{2,3\}$ and $\{3\}$ under the condition of not limiting the widths of spans.. The span $\{1, 2, 3\}$ is denoted as [1, 3]. We optimize $\theta$ to obtain not only the predicted answer $\hat{a}_{i}$ but also the span-level explanation $\hat{e}_i$ of $q_i$.

\section{Method}
In this section, we first given an overview of our method and then introduce the details of each module.
\label{method}
\subsection{Overview}
\begin{figure*}[htbp]
	\centering
	\includegraphics[width=\linewidth]{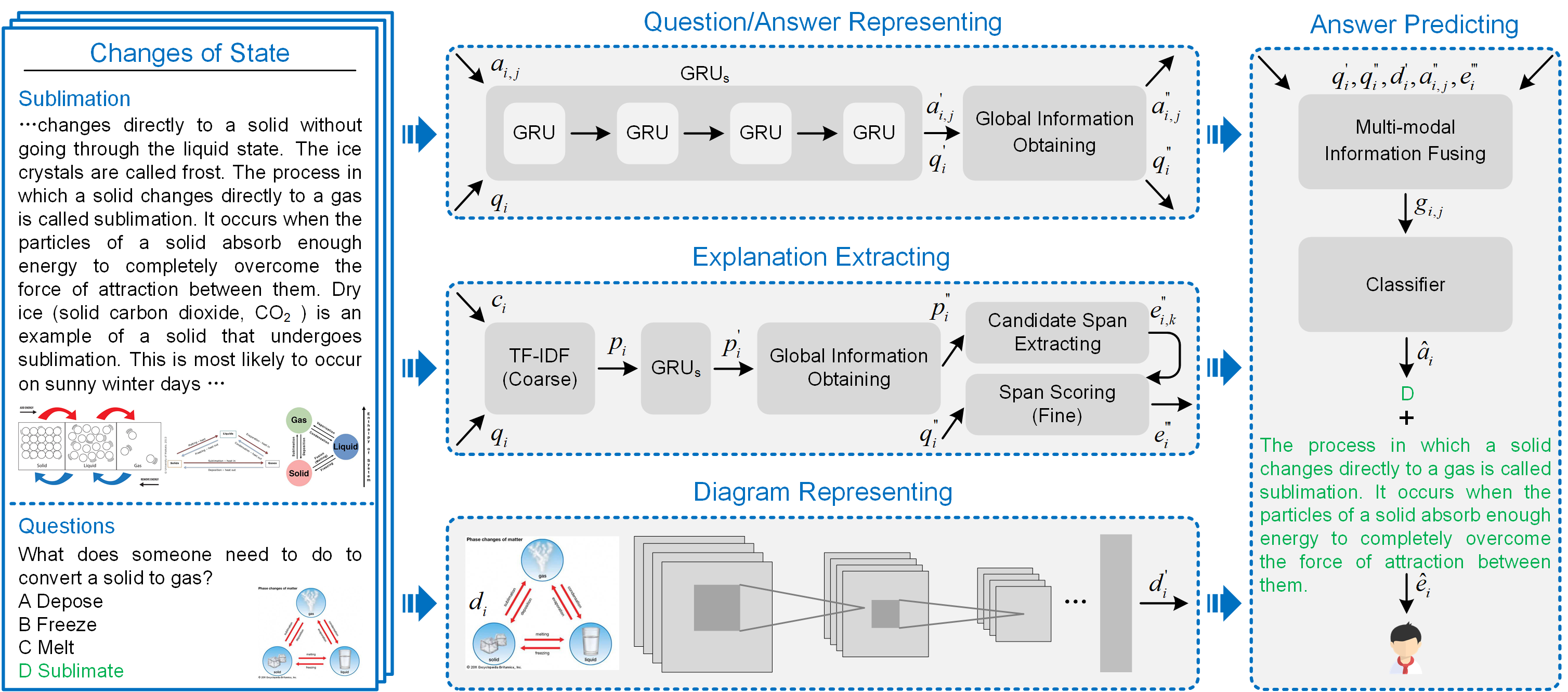}
	\caption{The architecture towards span-level eXplanations for Textbook Question Answering (XTQA). The left part shows a lesson and a question in the CK12-QA training split. The index of an evidence span is obtained by our proposed coarse-to-fine grained explanation extracting (EE) algorithm in the \emph{Explanation Extracting} module of the middle part. The \emph{Answer Predicting} module can not only provide answers but also the corresponding explanations for humans.}
	\label{fig.2}
\end{figure*}

The architecture of XTQA with four modules is shown in Figure \ref{fig.2}. XTQA first obtains the sentence-level representations $q_i^{''}, a_{i,j}^{''}$ of the question, candidate answer $q_i, a_{i,j}$ respectively in \emph{Question/Answer Representing}. Then, XTQA considers the entire textual context of a lesson as candidate evidence, and obtains the representations $e_{i}^{'''}$ of the top $U$ span-level explanations of $q_i$ and their indexes $[\mathrm{START}(k), \mathrm{END}(k)]$ using our proposed EE algorithm in \emph{Explanation Extracting}. Third, the contrastive learning is first used in the TQA task to learn effective diagram representations $d_i^{'}$ of $d_i$ in \emph{Diagram Representing}. Finally, XTQA gives humans not only the predicted answer $\hat{a}_{i}$ but also the span-level explanation $\hat{e}_i$ to choose it after fusing the above multi-modal information in \emph{Answer Predicting}. 

\subsection{Question/Answer Representing}
We use uni-directional Gated Recurrent Units (GRUs) to obtain the $r_1$-dimensional word-level representations $q_i^{'} \in \mathbb{R}^{X \times r_1}$, $a_{i,j}^{'} \in \mathbb{R}^{Y \times r_1}$ of $q_i$, $a_{i,j}$ respectively as follows:
\begin{equation}
\label{eq.1}
\begin{split}
& q_i^{'} = \mathrm{GRU_s}\big(embeding(q_i)\big), \\
& a_{i,j}^{'} = \mathrm{GRU_s}\big(embedding(a_{i,j})\big),
\end{split}
\end{equation}
where $q_i$ denotes the $i$-th question, $a_{i,j}$ denotes the $j$-th candidate answer of $q_i$, $X$ and $Y$ denote the maximum length of $q_i$ and $a_{i,j}$ respectively, and $embedding(\cdot)$ is used to learn the word embeddings.

To obtain the $r_1$-dimensional sentence-level representations $q_i^{''} \in \mathbb{R}^{r_1}$ of $q_i$, a learned attention mechanism is applied as follows:
\begin{equation}
\label{eq.2}
\begin{split}
& \alpha = \mathrm{softmax}\big(\mathrm{MLP_s}(q_i^{'})\big), \\
& q_i^{''} = \sum_{u=1}^{X} \alpha_u \circ q_{i,u}^{'},
\end{split}
\end{equation}
where $\alpha \in \mathbb{R}^{X}$ is the learned attention weight matrix by Multi-Layer Perceptrons ($\mathrm{MLP_s}$), $\circ$ denotes the element-wise product, and $q_{i,u}^{'} \in \mathbb{R}^{r_1}$ is the $u$-th word representations of $q_i$. 

Similarly, we also use the above learned attention mechanism to learn the $r_1$-dimensional sentence-level representations $a_{i,j}^{''} \in \mathbb{R}^{r_1}$ of $a_{i,j}$:
\begin{equation}
	\begin{split}
		& \alpha = \mathrm{softmax}\big(\mathrm{MLP_s}(a_{i,j}^{'})\big), \\
		& a_{i,j}^{''} = \sum_{u=1}^{Y} \alpha_u \circ a_{i,j,u}^{'},
	\end{split}
\end{equation}
where $a_{i,j,u} \in \mathbb{R}^{r_1}$ is the $u$-th word representations of $a_{i,j}$.

\subsection{Explanation Extracting}
Although the multi-modal context $c_i$ contains abundant essays with an average length of 788 words in CK12-QA, only a subset of sentences are required to answer $q_i$ and these sentences can also be regarded as the explanations for $q_i$. Inspired by this, XTQA first considers the entire textual context of a lesson as candidate evidence, and then extracts the top $U$ evidence spans from it as explanations using our proposed EE algorithm.

In the coarse-grained phase, we regard each paragraph of a lesson as a document and apply the embedding based query expansion method \cite{kuzi2016query} to narrow down the scope of textual contexts from a lesson to top $M$ paragraphs $p_i$ relevant to $q_i$. $p_i \in \mathbb{R}^{M \times L \times O}$ can be denoted as follows:
\begin{equation}
\label{eq.3}
p_i = \mathrm{Query}(q_i, c_i),
\end{equation}
where $L$ is the maximum number of sentences in each paragraph, and $O$ is the maximum length of each sentence. The shared $\mathrm{GRU_s}$ in Equation \ref{eq.1} is used to obtain the $r_1$-dimensional word-level representation $p_i^{'} \in \mathbb{R}^{M \times L \times O \times r_1}$ of $p_i$. We also use the shared learned attention mechanism in Equation \ref{eq.2} to obtain the $r_1$-dimensional sentence-level representations $p_i^{''} \in \mathbb{R}^{M \times L \times r_1}$ of $p_i$ to match the next phase.

In the fine-grained phase, the top $U$ span-level explanations are extracted from all candidate spans within top $M$ paragraphs by computing the information gain of each span for questions. Specifically, the representations at start $\mathrm{START}(k)$ and end $\mathrm{END}(k)$ indexes are concatenated to obtain the representation $e_{i,k}^{'} \in \mathbb{R}^{M \times N \times 2r_1}$ of the candidate evidence span $e_{i,k}$ as follows:
\begin{equation}
\label{eq.4}
e_{i,k}^{'} = \Big[p_{i, \mathrm{START}(k)}^{''}; p_{i, \mathrm{END}(k)}^{''}\Big],
\end{equation}
where $N = \frac{L(L+1)}{2}$ is the number of candidate evidence spans within each paragraph, $1 \leq k \leq N$, $1 \leq \mathrm{START}(k) \leq \mathrm{END}(k) \leq L$ and $[;]$ denotes the concatenation. To match the following steps, $e_{i,k}^{''} \in \mathbb{R}^{M \times N \times r_1}$ is obtained by the average pooling $\mathrm{AP}$ with kernel size $2 \times 1$ on $e_{i,k}^{'}$ as follows:
\begin{equation}
\label{eq.5}
e_{i,k}^{''} = \mathrm{AP}(e_{i,k}^{'}).
\end{equation}

XTQA computes the information gain $g(q_i, e_{i,k})$ of each candidate evidence span $e_{i,k}$ for $q_i$ to obtain the top $U$ span-level explanations. $g(q_i, e_{i,k})$ can be obtained as follows:
\begin{equation}
\label{eq.6}
g(q_i, e_{i,k}) = H(q_i) - H(q_i|e_{i,k}),
\end{equation}
where $H(q_i)$ is the entropy of $q_i$, \emph{i.e.}, the uncertainty of $q_i$, and $H(q_i|e_{i,k})$ is the conditional entropy of $q_i$ given $e_{i,k}$, \emph{i.e.}, the uncertainty of $q_i$ given $e_{i,k}$. The larger information gain indicates the more uncertainty of $q_i$ reduced by $e_{i,k}$. $H(q_i)$ can be obtained as follows:
\begin{equation}
\begin{split}
\label{eq.8}
& H(q_i) = \mathbb{E}\Big[-\log\big(p(q_i)\big)\Big], \\
& p(q_i) = \sigma\big(\mathrm{MLP_h}(q_i^{''})\big),
\end{split}
\end{equation}
where $\mathbb{E}$ is the expected value operator, $p(q_i)$ denotes the probability of $q_i$ being answered accurately, $q_i^{''}$ denotes the sentence-level representations of $q_i$ and $\sigma$ is the sigmoid function. $H(q_i|e_{i,k})$ can be obtained as follows: 
\begin{equation}
\begin{split}
\label{eq.9}
& H(q_i|e_{i,k}) = \mathbb{E}\Big[-\log\big(p(q_i, e_{i,k})\big)\Big], \\
& p(q_i, e_{i,k}) = \sigma\bigg(\mathrm{MLP_h}\Big(\mathrm{AP}\big([q_i^{''}, e_{i,k}^{''}]\big)\Big)\bigg),
\end{split}
\end{equation}
where $p(q_i, e_{i,k})$ is the probability of $q_i$ being answered accurately given $e_{i,k}$ and $\mathrm{AP}$ is the average pooling with kernel size $2 \times 1$. A formal description about the algorithm is shown in Algorithm \ref{algorithm1}.

After obtaining the top $U$ span-level explanations and their representations $e_i^{''} \in \mathbb{R}^{U \times r_1}$, the learned attention mechanism in Equation \ref{eq.2} is used to obtain the global span-level explanation representation $e_{i}^{'''} \in \mathbb{R}^{r_1}$.
\begin{algorithm}[tbp]
	\caption{Explanation extracting}
	\label{algorithm1}
	\begin{algorithmic}[1]
		\Require
		question $q_i$, multi-modal context $c_i$. 
		\Ensure 
		representation $e_i^{''}$ of span-level explanation $e_i$ and its index. 
		\State Choose top $M$ paragraphs relevant to $q_i$ using Equation \ref{eq.3}; 
		\State Construct the possible span according to the way described in Section \ref{task-for}; 
		\State Obtain the candidate evidence span representation using Equation \ref{eq.4};
		\State Obtain the global representation of the span using Equation \ref{eq.5}; 
		\State Compute the entropy of $q_i$ using Equation \ref{eq.8}; 
		\State Compute the conditional entropy of $q_i$ given each candidate evidence span $e_{i,k}$ using Equation \ref{eq.9}; 
		\State Compute the information gain of each span for $q_i$ using Equation \ref{eq.6}; 
		\State Select top $U$ span-level explanations according to the gain.
	\end{algorithmic}
\end{algorithm}

\subsection{Diagram Representing}
Effective diagram representations play a key role in improving the TQA performance. However, there has no annotation for diagrams in CK12-QA. Recently, self-supervised learning methods such as SimCLR \citep{chen2020simple} have made significant progress in image classification, which shows they can learn the deep understanding of images. Inspired by this, we first pre-train CNNs such as ResNet on the diagrams within CK12-QA by contrastive learning \citep{chen2020simple} and fine-tune this module on the TQA task to learn the $r_2$-dimensional representation $d_i^{'} \in \mathbb{R}^{r_2}$ of $d_i$ as follows:
\begin{equation}
\label{d_i}
d_i^{'} = \mathrm{CNNs}(d_i).
\end{equation}
\subsection{Answer Predicting}
After the above modules' processing, we obtain the word and sentence-level representations $q_i^{'}, q_i^{''}$ of $q_i$, the diagram representation $d_i^{'}$ of $d_i$, the sentence-level representation $a_{i,j}^{''}$ of $a_{i,j}$, the span-level explanation representation $e_i^{'''}$ of $e_{i}$ and the indexes $[\mathrm{START}(k), \mathrm{END}(k)]$ of spans. In general, multi-grained or multi-level representations are beneficial for obtaining effective multi-modality features. Therefore, they are used to obtain the global fusion feature $g_{i,j} \in \mathbb{R}^{9r_1}$ with $j$-th candidate answer as follows:
\begin{equation}
\label{multimodal}
\begin{split}
& g_{i,j} = \big[q_i^{''}; d_i^{'}; a_{i,j}^{''}; e_i^{'''}; g_{i}^{\beta}; g_{i,j}^{\gamma}; g_{i}^{\mu}; g_{i,j}^{\eta}; g_{i,j}^{\psi}\big], \\
& g_{i}^{\beta} = \mathrm{BAN}(q_i^{'}, d_i^{'}), ~ g_{i,j}^{\gamma} = Wq_i^{''} \circ Wa_{i,j}^{''}, \\
& g_{i}^{\mu} = Wq_i^{''} \circ We_i^{'''}, ~ g_{i,j}^{\eta} = We_i^{'''} \circ Wa_{i,j}^{''},    \\
& g_{i,j}^{\psi} = Wq_i^{''} \circ Wa_{i,j}^{''} \circ g_{i}^{\beta},
\end{split}
\end{equation}
where $\mathrm{BAN}$ is the bi-linear attention mechanism \citep{kim2018bilinear}, $W \in \mathbb{R}^{r_1 \times r_1}$ is the learned weight matrix, $g_{i}^{\beta}, g_{i,j}^{\gamma}, g_{i}^{\mu}, g_{i,j}^{\eta} \in \mathbb{R}^{r_1}$ denote the pairwise similarity, and $g_{i,j}^{\psi} \in \mathbb{R}^{r_1}$ is the triple-wise similarity.

To obtain the scores of candidate answers $s_i \in \mathbb{R}^{|\mathcal{A}_i|}$, $g_{i} \in \mathbb{R}^{|\mathcal{A}_i| \times 9r_1}$ is projected as follows:
\begin{equation}
s_i = \mathrm{MLP_c}(g_{i}),
\end{equation}
where $|\mathcal{A}_i|$ denotes the number of candidate answers of $q_i$.

We regard the TQA task as a multi-class classification. Therefore, XTQA is optimized by the multi-class cross-entropy function as follows:
\begin{equation}
\begin{split}
& \mathcal{L} = -\sum_{i=1}^{n} y_i\log \hat{y}_i, \\
& \hat{y}_i = \mathrm{softmax}(s_i),
\end{split}
\end{equation}
where $n$ denotes the number of questions, $y_i \in \{0, 1\}^{|\mathcal{A}_i|}$ denotes the true answers of $q_i$, $\hat{y}_i \in [0,1]^{|\mathcal{A}_i|}$ denotes the predicted probability of candidate answers and $\mathrm{softmax}$ denotes the softmax function.

Eventually, not only the predicted answer $\hat{a}_{i}$ but also the span-level explanation $e_i$ is provided for humans.

\section{Experiments}
\label{experiment}
In this section, we first introduce the experimental setups such as datasets and implement details. Then, we describe the results including explanation and TQA accuracy. Third, ablation studies and discussions are introduced. Finally, we give a case study of XTQA. 

\subsection{Datasets and Evaluations}
Currently, there exist two TQA datasets including CK12-QA \cite{kembhavi2017you} and AI2D \cite{kembhavi2016diagram}. Most of the previous works \cite{li2018textbook, kim2019textbook, HauriletAS18} are only evaluated on CK12-QA except ISAAQ \cite{gomez2020isaaq}. Due to the lack of multi-modal contexts of AI2D, XTQA can not be applicable to this dataset to extract explanations. We will explore how to generate explanations only given questions and diagrams in the future. Following the previous works, we evaluate XTQA on CK12-QA \citep{kembhavi2017you} that consists of $1,076$ lessons with $78,338$ sentences and $3,455$ diagrams. The lessons are obtained from the \emph{physical science}, \emph{life science} and \emph{earth science} textbooks of the middle school on-line curricula. The dataset is split into a training set with $666$ lessons and $15,154$ questions, a validation dataset with $200$ lessons and $5,309$ questions, and a test set with $210$ lessons and $5,797$ questions. Among the total $26,260$ questions, $12,567$ of them have an accompanying diagram. There are four candidate answers for each diagram question. The non-diagram questions can be classified into two categories: True/False (T/F) with two candidate answers and Multiple Choice (MC) with four to seven candidate answers.

To estimate the results of span-level explanations, we employ the mean Intersection over Union (mIoU) that is always used in object detection \cite{zhao2019object} and segmentation \cite{zhu2019improving} as metrics. The TQA accuracy is obtained by checking whether the prediction is the same as the ground truth.

\subsection{Implementation Details}
In \emph{Question/Answer Representing}, we use BERT \citep{devlin2019bert} to obtain $768$-dimensional word embeddings and apply uni-directional one-layer $\mathrm{GRUs}$ with $r_1=1024$ hidden units to encode questions and candidate answers. The shared $\mathrm{MLPs}$ (FC($1024$)-Dropout($0.2$)-FC($1$)) is used to learn attention coefficients. In \emph{Explanation Extracting}, the pylucene is used to conduct paragraphs indexing and searching. The maximum number of paragraphs $M$, the maximum number of sentences within each paragraph $L$, the maximum length of each sentence $O$ and the maximum number of span-level explanations $U$ are set to $1/1, 5/15, 20/15$ and $1/1$ for non-diagram/diagram question answering respectively. We set the maximum widths $W$ of candidate evidence spans to $2$. In \emph{Diagram Representing}, we resize the diagrams to $224$ owing to the different sizes of them in the dataset. To obtain $r_2=2048$-dimensional diagram representations, we first train the SimCLR on the diagrams in CK12-QA with default hyper-parameters, and then fine-tune the pre-trained model by the task-specific supervision (TQA). In \emph{Answer Predicting}, the $\mathrm{MLPc}$ (FC($2048$)-ReLU-Dropout($0.2$)-FC($1$)) is used to obtain the candidate answer scores.

XTQA is trained by the Adam optimizer with $\beta_1=0.9, \beta_2=0.98$. The base learning rate is $\min(2.5\tau e^{-4}, 1e^{-4})$, where $\tau$ is the current epoch. The rate is decayed by $0.1$ after $8$ epochs. XTQA converges at the end of the $10$-th epoch with the batch size $2$. Parameters of XTQA are initialized by the Pytorch default initialization with the fixed seed $666$. All the experiments are run on one NVIDIA’s Tesla V100 card.
\subsection{Explanation Results}
\label{explanation}
To the best of our knowledge, XTQA is the first method to explore TQA explanations. We select five previous state-of-the-art methods which focus on multi-modality fusion and match well with our proposed EE algorithm as baselines. The introductions of them are as follows.

\begin{itemize}
	\item[1)] MFB \citep{yu2017multi} is a multi-modal factorized bi-linear pooling approach, which aims at addressing the high dimensionality of the output features and the huge number of parameters caused by bi-linear pooling based models \citep{tenenbaum1997}.
	
	\item[2)] MUTAN \citep{ben2017mutan} is a multi-modal tensor-based decomposition approach with a low-rank matrix constraint. It also aims at addressing the huge dimensionality issue.
	
	\item[3)] BAN \citep{kim2018bilinear} is a bi-linear attention network that aims at learning effective interactions between images and questions using the proposed bi-linear attention mechanism.
	
	\item[4)] MCAN \citep{yu2019deep} is a deep modular co-attention network. It aims at obtaining sufficient multi-modality interactions by modularly composing the self-attention of questions and images, as well as the question-guided-attention of images.
	
	\item[5)] CMR \cite{ZhengGK20} is a cross-modality relevance network, which learns the relevance representations between entities of input modalities and models the higher-order relevance between entity relations to perform language and vision reasoning. It is the current state-of-the-art method for VQA.
\end{itemize}

To explore how well XTQA and \emph{baselines}+EE provide explanations for humans, we manually annotate span-level explanations for the first $150$ non-diagram and diagram questions of the validation and test split respectively. Note that we do not use these annotations to train XTQA and \emph{baselines} + EE. We apply mIoU to evaluate their performance. For example, if the indexes of a predicted span are (3, 5) and the indexes of its corresponding gold span are (4, 5), the IoU value is 0.67. We do not consider whether the question is accurately answered here.
\begin{table}[tbp]
	\centering
	\caption{The explanation results (\%) on the first 150 Non-Diagram (ND) and 150 diagram questions within the validation split.}
	\label{tab-val-exp}
	\resizebox{\columnwidth}{!}{	
		\begin{threeparttable}
			\begin{tabular}{lccccc}
				\toprule[1pt]
				Model     						& ND T/F 		  & ND MC  			 & ND All 		    & Diagram 		     & All    		\\ \midrule[1pt]
				MFB   \cite{yu2017multi}+EE 	& $50.96$     	  & $40.25$          & $44.46$          & $45.68$   	     & $45.07$		\\
				MUTAN \cite{ben2017mutan}+EE	& $52.36$     	  & $43.39$ 		 & $46.92$          & $43.36$   	     & $45.14$  	\\
				BAN   \cite{kim2018bilinear}+EE & $56.38$     	  & $42.66$   	     & $\textbf{48.06}$ & $47.20$   	     & $47.63$		\\
				MCAN  \cite{yu2019deep}+EE	 	& $53.21$     	  & $\textbf{43.58}$ & $47.37$		    & $48.69$   	     & $48.03$		\\ 
				CMR   \cite{ZhengGK20}+EE 		& $51.82$     	  & $41.20$   	     & $45.38$          & $51.23$            & $48.30$		\\ \midrule[1pt]
				XTQA      						& $\textbf{58.69}$& $40.59$ 		 & $47.71$ 			& $\textbf{52.41}$   & $\textbf{50.06}$  \\ \bottomrule[1pt]
			\end{tabular}
			\begin{tablenotes}
				\item[1] EE denotes our proposed explanation extracting algorithm.
				\item[2] ND All = ND T/F $\cup$ ND MC and All = ND All $\cup$ Diagram.
			\end{tablenotes}
	\end{threeparttable}}
\end{table}
\begin{table}[tbp]
	\centering
	\caption{The explanation results (\%) on the first 150 Non-Diagram (ND) and 150 diagram questions within the test split.}
	\label{tab-test-exp}
	\resizebox{\columnwidth}{!}{	
			\begin{tabular}{lccccc}
				\toprule[1pt]
				Model     						& ND T/F 		  & ND MC  			 & ND All 		    & Diagram 		     & All    		\\ \midrule[1pt]
				MFB   \cite{yu2017multi}+EE 	& $44.02$     	  & $46.98$          & $45.93$          & $51.05$   	     & $48.49$		\\
				MUTAN \cite{ben2017mutan}+EE	& $43.89$     	  & $48.52$   	     & $46.88$          & $51.35$   	     & $49.12$  	\\
				MCAN  \cite{yu2019deep}+EE	 	& $43.88$     	  & $\textbf{51.89}$ & $49.06$          & $51.56$   	     & $50.31$		\\ 
				CMR   \cite{ZhengGK20}+EE 		& $50.11$     	  & $47.25$   	     & $48.26$          & $\textbf{52.86}$   & $50.56$		\\
				BAN   \cite{kim2018bilinear}+EE & $51.29$     	  & $49.69$   	     & $50.26$          & $52.33$   	     & $51.29$		\\ \midrule[1pt]
				XTQA      						& $\textbf{55.75}$& $49.88$ 		 & $\textbf{51.95}$ & $52.80$            & $\textbf{52.38}$  \\ \bottomrule[1pt]
			\end{tabular}}
\end{table}

Table \ref{tab-val-exp} and \ref{tab-test-exp} show the results on the validation and test split respectively. It can be seen that each method provides more accurate explanations for T/F questions compared with the results on MC questions within the validation split. However, the results do not have the same trend on the test split. This may be caused by the different data distributions between the validation and test split. For example, the knowledge of lesson \emph{earth science and its branches} is mutually exclusive between train and validation split. In addition, we can also see that \emph{baselines}+EE also have the ability to provide explanations for humans. XTQA, MCAN \cite{yu2019deep}+EE and CMR \cite{ZhengGK20}+EE obtain the best explanations for T/F, ND MC and diagram questions on the test split respectively. In brief, XTQA obtains the best result of explanations on the validation and test split compared with \emph{baselines}+EE and the EE algorithm can be integrated into other TQA methods to make them explainable.
\subsection{TQA Accuracy}
To explore how much the TQA performance that XTQA, baselines and \emph{baselines}+EE can obtain, we conduct experiments on the validation and test split of CK12-QA. The results are shown in Table \ref{tab.1} and \ref{tab.2} respectively.

\begin{table}[tbp]
	\centering
	\caption{The TQA accuracy (\%) on different-type questions within the validation split.}
	\label{tab.1}
	\resizebox{\columnwidth}{!}{	
		\begin{threeparttable}
			\begin{tabular}{lccccc}
				\toprule[1pt]
				Model     						& ND T/F 		  & ND MC  			 & ND All 		    & Diagram 		     & All    		\\ \midrule[1pt]
				CMR   \cite{ZhengGK20} 			& $51.14$     	  & $30.65$   	     & $38.72$          & $30.73$   	     & $34.53$		\\
				MUTAN \cite{ben2017mutan}		& $51.72$     	  & $31.18$   	     & $39.27$          & $30.29$   	     & $34.56$  	\\
				MFB   \cite{yu2017multi} 		& $51.73$     	  & $30.65$          & $38.95$          & $30.76$   	     & $34.65$		\\
				BAN   \cite{kim2018bilinear} 	& $51.70$     	  & $31.11$   	     & $39.22$          & $30.65$   	     & $34.73$		\\
				MCAN  \cite{yu2019deep}	 		& $51.72$     	  & $32.55$ 		 & $40.10$          & $30.58$   	     & $35.10$		\\ \hdashline
				MFB \cite{yu2017multi}+EE 	    & $52.82$     	  & $31.76$   	     & $40.05$          & $31.97$   	     & $35.82$		\\
				MUTAN \cite{ben2017mutan}+EE 	& $53.02$     	  & $32.55$   	     & $40.61$          & $31.75$   	     & $35.97$		\\
				BAN \cite{kim2018bilinear}+EE   & $54.73$     	  & $31.44$   	     & $40.61$          & $31.82$   	     & $36.00$		\\
				CMR \cite{ZhengGK20}+EE 		& $52.11$     	  & $31.83$   	     & $39.82$          & $\textbf{32.87}$   & $36.18$		\\
				MCAN \cite{yu2019deep}+EE 	    & $52.92$     	  & $\textbf{32.88}$ & $40.77$          & $32.22$   	     & $36.29$		\\ \midrule[1pt]
				XTQA      						& $\textbf{58.24}$& $30.33$ 		 & $\textbf{41.32}$ & $32.05$   & $\textbf{36.46}$  \\ 
				\bottomrule[1pt]
			\end{tabular}
			\begin{tablenotes}
				\item[1] EE denotes our proposed explanation extracting algorithm.
				\item[2] ND All = ND T/F $\cup$ ND MC and All = ND All $\cup$ Diagram.
				\item[3] All the baselines use the same BERT embeddings and diagram representations as XTQA for a fair comparison.
			\end{tablenotes}
	\end{threeparttable}}
\end{table}
\begin{table}[tbp]
	\centering
	\caption{The TQA accuracy (\%) on different-type questions within the test split.}
	\label{tab.2}
	\resizebox{\columnwidth}{!}{	
			\begin{tabular}{lccccc}
				\toprule[1pt]
				Model     								& ND T/F 		  & ND MC  			 & ND All 		    & Diagram 		     & All    		\\ \midrule[1pt]
				BAN \cite{kim2018bilinear} 			    & $48.08$     	  & $32.96$   	     & $38.44$          & $27.28$   	     & $32.11$		\\
				MFB   \cite{yu2017multi} 				& $48.08$     	  & $31.83$ 		 & $37.72$          & $28.17$   	     & $32.30$		\\
				MCAN  \cite{yu2019deep}	 			    & $48.08$     	  & $33.15$   	     & $38.56$          & $27.56$   	     & $32.32$		\\ 
				MUTAN \cite{ben2017mutan}				& $48.08$     	  & $32.90$   	     & $38.40$          & $28.29$   	     & $32.67$  	\\
				CMR   \cite{ZhengGK20} 					& $52.02$     	  & $33.15$   	     & $39.99$          & $29.54$   	     & $34.06$		\\ \hdashline
				MUTAN \cite{ben2017mutan}+EE 			& $50.17$     	  & $33.21$   	     & $39.36$          & $32.96$   	     & $35.73$		\\
				MFB \cite{yu2017multi}+EE 			    & $51.70$     	  & $32.77$   	     & $39.63$          & $32.90$   	     & $35.81$		\\
				MCAN \cite{yu2019deep}+EE 				& $51.05$     	  & $\textbf{34.08}$ & $40.23$          & $33.21$   	     & $36.25$		\\
				CMR \cite{ZhengGK20}+EE 				& $54.13$     	  & $32.15$   	     & $40.12$          & $\textbf{33.40}$   & $36.31$		\\ 
				BAN \cite{kim2018bilinear}+EE 			& $53.58$     	  & $33.33$   	     & $40.67$          & $33.15$   	     & $36.41$\\ \midrule[1pt]
				XTQA 		     						& $\textbf{56.22}$& $33.40$& $\textbf{41.67}$ 		    & $33.34$   & $\textbf{36.95}$  \\ 
				\bottomrule[1pt]
			\end{tabular}}
\end{table}
In Table \ref{tab.1}, we can see that XTQA outperforms the best baseline MCAN \cite{yu2019deep} and the best \emph{baselines}+EE MCAN+EE by $1.36\%$ and $0.17\%$ on the total questions of the validation split respectively. For diagram questions, XTQA outperforms the best baseline MFB \cite{yu2017multi} by $1.29\%$ and CMR \cite{ZhengGK20}+EE achieves the best results. For non-diagram questions, XTQA outperforms the best baseline MCAN and the best \emph{baselines}+EE MCAN+EE by $1.22\%$ and $0.55\%$ respectively. For T/F questions, XTQA outperforms the best baseline MFB and the best \emph{baselines}+EE BAN \cite{kim2018bilinear}+EE by $6.51\%$ and $3.51\%$. XTQA has the worst performance on the MC questions, which may be caused by the different data distributions between T/F (two candidate answers) and MC (four to seven candidate answers) questions. In this paper, we regard both of the ND T/F and ND MC subtasks as the multi-class classification, \emph{i.e.}, padding the number of candidate answers of ND T/F questions into seven following previous works \cite{li2018textbook}. In this way, we do not need to devise a specific model for ND T/F and ND MC respectively. Note that all the baselines perform information fusion between the top $1$ paragraph and questions for non-diagram questions considering the specificity of the TQA task.

In Table \ref{tab.2}, we can see that XTQA outperforms the best baseline CMR \cite{ZhengGK20} and the best \emph{baselines}+EE BAN \cite{kim2018bilinear}+EE by $2.89\%$ and $0.54\%$ on the total questions of the test split respectively. For diagram questions, XTQA outperforms the best baseline CMR by $3.80\%$ and CMR+EE achieves the best result. For non-diagram questions, XTQA is superior to the best \emph{baselines}+EE BAN+EE by $1\%$ and achieves the best accuracy. For TF questions, XTQA outperforms the best baseline CMR and the best \emph{baselines}+EE CMR+EE by $4.20\%$ and $2.09\%$ respectively. For MC questions, XTQA outperforms the best baseline MCAN \cite{yu2019deep} and CMR by $0.25\%$ and MCAN+EE achieves the best result. In short, XTQA achieves the best performance on the two splits, which shows the effectiveness of our method. In addition, the results also demonstrate EE can not only enhance the explainability of baselines but also the TQA performance of them.

To the best of our knowledge, existing TQA methods except ISAAQ and RAFR lack the results on the CK12-QA test split. ISAAQ achieves the current state-of-the-art results based on large datasets pre-training, large pre-trained model fine-tuning and ensemble learning. RAFR achieves modest results with training only on CK12-QA and without large pre-trained model fine-tuning and ensemble learning. To fairly compare with ISAAQ, we make the following minor changes for our method as follows.
\begin{itemize}
	\item[1)] Following ISAAQ, we apply RoBERTa to obtain the word-level representations $q_i^{'} \in \mathbb{R}^{X \times r_1}$ of $q_i$, which is similar to Equation \ref{eq.1}. The word-level representation $p_i^{'} \in \mathbb{R}^{M \times L \times O \times r_1}$ of $p_i$ is also learned by RoBERTa.
	\item[2)] We concatenate $q_i$ and $a_{i,j}$, and apply RoBERTa to learn a joint sentence-level representation $qa_{i,j} \in \mathbb{R}^{r_1}$, which is different from Equation \ref{eq.2}.
	\item[3)] We do not use the multi-modal fusing in Equation \ref{multimodal}. A question-explanation guided gate mechanism is proposed to learn the attended diagram representation $d_{i,j}^{''} \in \mathbb{R}^{r_1}$ of $qa_{i,j}$. Then, we concatenate them to obtain the fusion information $g_{i,j} \in \mathbb{R}^{r_1}$. The above steps can be denoted as follows:
	\begin{equation}
		\begin{split}
			& \alpha_{i,j} = \sigma\Big(W_a\big([qa_{i,j}; e_i^{'''}]\big)\Big), \\
			& d_{i,j}^{''} = \alpha_{i,j} \odot (W_d d_i^{'}), \\
			& g_{i,j} = [qa_{i,j}; e_i^{'''}; d_{i,j}^{''}],
		\end{split}
	\end{equation}
	where $\sigma$ is the sigmoid function, $W_a \in \mathbb{R}^{r_1 \times 2r_1}$ and $W_d \in \mathbb{R}^{r_1 \times r_2}$ are learned weight matrices, $\alpha_{i,j}$ denotes the gate weight, $d_i^{'}$ is the diagram representation learned by Equation \ref{d_i} and $e_i^{'''}$ denotes the explanation representation obtained by Algorithm \ref{algorithm1}.
\end{itemize}

We call this method XTQA-V2. We train it and ISAAQ only on CK12-QA to make a fair comparison. ISAAQ \cite{gomez2020isaaq} employs information retrieval, next sentence prediction, and nearest neighbors to extract the most related paragraphs of questions respectively. Therefore, we use the average accuracy as the final result. The results on the validation and test splits are shown in Table \ref{tab-xtqav2-val} and \ref{tab-xtqav2} respectively. We can see that RoBERTa-based models significantly outperform XTQA and RAFR that do not use the large language model fine-tuning. This is caused by large parameters and prior transfer knowledge.
\begin{table}[tbp]
	\centering
	\caption{The comparison with current state-of-the-art TQA methods on the validation split.}
	\label{tab-xtqav2-val}
	\resizebox{\columnwidth}{!}{	
		\begin{threeparttable}
			\begin{tabular}{clccccc}
				\toprule[1pt]
				LMF & Model     				& ND T/F 		  & ND MC  			 & ND All 		    & Diagram 		   & All   	\\ \midrule[1pt]
				No	& RAFR \cite{ma2021relation}& $53.63$ 		  & $\textbf{36.67}$ & $\textbf{43.35}$          & $\textbf{32.85}$ & $\textbf{37.85}$  \\ 
				No	& XTQA 						& $\textbf{58.24}$& $30.33$          & $41.32$ & $32.05$          & $36.46$  \\ \hdashline
				Yes	& ISAAQ\textsubscript{avg} \cite{gomez2020isaaq}& $72.67$        & $54.77$          & $61.84$          & $39.13$          & $49.94$  \\
				Yes	& XTQA-V2		     		& $\textbf{76.65}$& $\textbf{57.65}$ & $\textbf{65.15}$ & $\textbf{46.85}$ & $\textbf{55.56}$  \\ 
				\bottomrule[1pt]
			\end{tabular}
			\begin{tablenotes}
				\item[1] LMF denotes whether a specific method uses pre-trained language model fine-tuning. ND All = ND T/F $\cup$ ND MC and All = ND All $\cup$ Diagram.
				\item[2] RAFR and XTQA regards ND T/F and ND MC as a multi-class classification, which means the non-diagram questions does not be distinguished manually. XTQA-V2 and ISAAQ regards ND T/F as a binary classification and ND MC as a multi-class classification.
			\end{tablenotes}
	\end{threeparttable}}
\end{table}
\begin{table}[tbp]
	\centering
	\caption{The comparison with current state-of-the-art TQA methods on the test split.}
	\label{tab-xtqav2}
	\resizebox{\columnwidth}{!}{	
			\begin{tabular}{clccccc}
				\toprule[1pt]
				LMF & Model     				& ND T/F 		  & ND MC  			 & ND All 		    & Diagram 		   & All   	\\ \midrule[1pt]
				No	& RAFR \cite{ma2021relation}& $52.75$ 		  & $\textbf{34.38}$ & $41.03$          & $30.47$ 		   & $35.04$  \\ 
				No	& XTQA 						& $\textbf{56.22}$& $33.40$          & $\textbf{41.67}$ & $\textbf{33.34}$ & $\textbf{36.95}$  \\ \hdashline
				Yes	& ISAAQ\textsubscript{avg} \cite{gomez2020isaaq}& $72.62$        & $55.94$          & $62.00$          & $35.18$          & $46.80$  \\
				Yes	& XTQA-V2		     		& $\textbf{75.88}$& $\textbf{61.56}$ & $\textbf{66.76}$ & $\textbf{41.04}$ & $\textbf{52.16}$  \\ 
				\bottomrule[1pt]
			\end{tabular}}
\end{table}

\subsection{Ablation Studies}
We perform ablation studies on the validation split shown in Table \ref{tab.3} to analyze the effectiveness of each module. 
\begin{table}[!htbp]
	\centering
	\caption{Ablation results ($\%$ accuracy) on the validation split. $\Delta$ denotes the accuracy reduction without the specific module.}
	\label{tab.3}
	\resizebox{\columnwidth}{!}{	
		\begin{threeparttable}
			\begin{tabular}{lcccccc}
				\toprule[1pt]
				Models                               & ND All            & $\Delta$ 		& Diagram 			 & $\Delta$         & All   			& $\Delta$     \\ \midrule[1pt]
				XTQA                           		 & $\mathbf{41.32}$  &  				& $\mathbf{32.05}$   &  			    & $\mathbf{36.46}$  &  	           \\ 
				\hdashline
				w/o contrastive learning             & $-$  			 & $-$ 		    	& $30.03$       	 & $-2.03$	        & $35.41$			& $-1.05$	   \\
				w/o fine-tuning ResNet               & $-$  			 & $-$ 		    	& $30.36$       	 & $-1.69$	        & $35.57$			& $-0.89$	   \\
				w/o BERT embedding                   & $41.20$           & $-0.12$      	& $29.82$        	 & $\mathbf{-2.23}$ & $35.24$      		& $-1.22$ 	   \\ 
				w/o span-level explanation           & $38.95$           & $\mathbf{-2.37}$ & $30.36$        	 & $-1.69$ 		    & $34.45$           & $\mathbf{-2.01}$ 	   \\ \bottomrule[1pt]
			\end{tabular}
			\begin{tablenotes}
				\item[1] ND All = ND T/F $\cup$ ND MC and All = ND All $\cup$ Diagram.
			\end{tablenotes}
		\end{threeparttable}
	}
\end{table}
\begin{table}[!htbp]
	\centering
	\caption{Algorithm analysis ($\%$ accuracy) on the validation split. $\Delta$ denotes the accuracy reduction.}
	\label{tab: algorithm-an }	
	\begin{threeparttable}
		\begin{tabular}{cccccc}
			\toprule[1pt]
			$M$  & $U$ & $W$   	& ND All           & Diagram          & All   			 \\ \midrule[1pt]
			$1$  & $-$ & $-$    & $38.95$          & $30.36$          & $34.45$           \\
			$1$  & $1$ & $1$    & $40.52$          & $31.85$          & $35.97$   \\  
			$1$  & $1$ & $2$    & $41.32$          & $\mathbf{32.05}$ & $\mathbf{36.46}$   \\ 
			$1$  & $2$ & $2$    & $41.47$          & $31.60$          & $36.30$   \\  
			$1$  & $1$ & $3$    & $\mathbf{41.59}$ & $31.72$          & $36.41$           \\ \hdashline
			$2$  & $-$ & $-$    & $39.01$          & $30.08$          & $34.33$           \\ 
			$2$  & $1$ & $1$    & $40.99$          & $31.34$          & $35.93$           \\ 
			$2$  & $1$ & $2$    & $41.41$          & $31.78$          & $36.37$           \\  
			$2$  & $2$ & $2$    & $41.35$          & $31.38$          & $36.13$           \\  
			$2$  & $2$ & $3$    & $41.06$          & $31.25$          & $35.92$           \\  
			\bottomrule[1pt]
		\end{tabular}
		\begin{tablenotes}
			\item[1] Top $M$ paragraphs, top $U$ spans in each paragraph and span width with $W$.
		\end{tablenotes}
	\end{threeparttable}
\end{table}

\noindent\textbf{W/O Contrastive Learning} XTQA does not use the contrastive learning to pre-train ResNet but uses the answer label to fine-tune it in the experiment. The accuracy on the diagram questions drops by $2.03\%$, which demonstrates the contrastive learning can be used to learn effective diagram representations except learning image representations.

\noindent\textbf{W/O Fine-tuning ResNet} The ResNet is pre-trained by contrastive learning but not fine-tuned using the answer label in the experiment. The accuracy on the diagram questions drops by $1.69\%$, which shows that effective diagram representation is important to improve the TQA performance. 

\noindent\textbf{W/O BERT Embedding} XTQA does not apply BERT embeddings to initialize the weight of embedding layers and applies the PyTorch default initialization in the experiment. The accuracy on the non-diagram, diagram questions drop by $0.12\%$ and $2.23\%$ respectively, which shows the significant performance difference of the BERT embeddings on the different task. This may be caused by unstable optimization on the diagram question answering because there exists a large vocabulary with 19556 words but few training data with $6,501$ diagram questions.

\noindent\textbf{W/O Span-level Explanation} XTQA does not apply the fine-grained explanation but uses the coarse-grained top $M$ paragraphs in the experiment. The accuracy on non-diagram, diagram questions drop by $2.37\%$ and $1.69\%$ respectively, which shows the importance of span-level explanations. Moreover, the accuracy on the whole questions decreases the most compared with the ablation study of other modules.

In short, each component makes its contributions to the performance of XTQA and our proposed coarse-to-fine grained EE algorithm plays the biggest role.

\subsection{Algorithm Analyses}
We conduct the experiments shown in Table \ref{tab: algorithm-an } on the validation split to explore the effect of $M$, $U$, and span width of Algorithm \ref{algorithm1} on TQA accuracy. All the other hyper-parameters are the same in these experiments. It is clear that choosing the top $M=1$ paragraph is the best way to answer non-diagram and diagram questions. Our method performs better on non-diagram questions but worse on diagram questions as the number of spans increases. The preceding description also applies to span width. We can conclude that the long textual explanations would interfere with answering diagram questions.
\begin{figure}[tbp]
	\centering
	\includegraphics[width=0.7\linewidth]{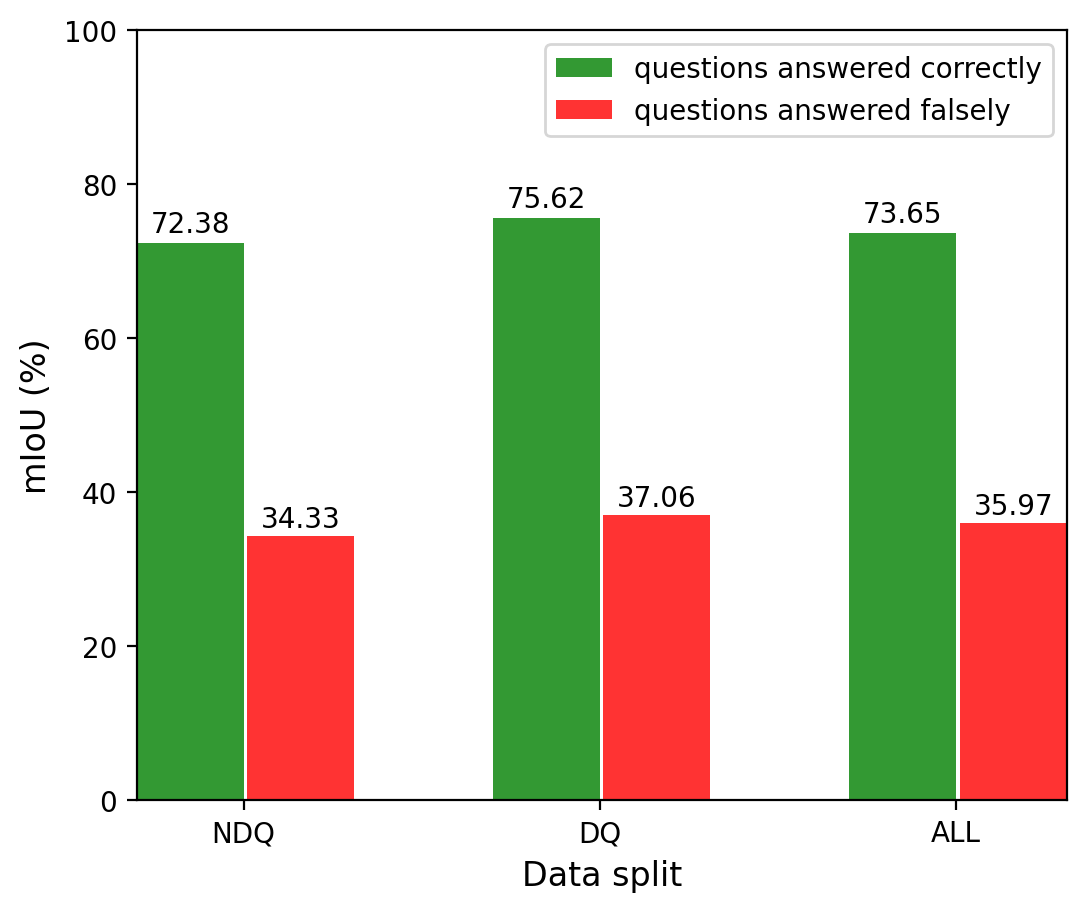}
	\caption{The explanation result (\%) of XTQA under the condition of answering questions correctly and incorrectly. These questions are introduced in Section \ref{explanation}.}
	\label{fig-span-acc}
\end{figure}
\begin{figure}[!tbp]
	\centering
	\includegraphics[width=0.7\linewidth]{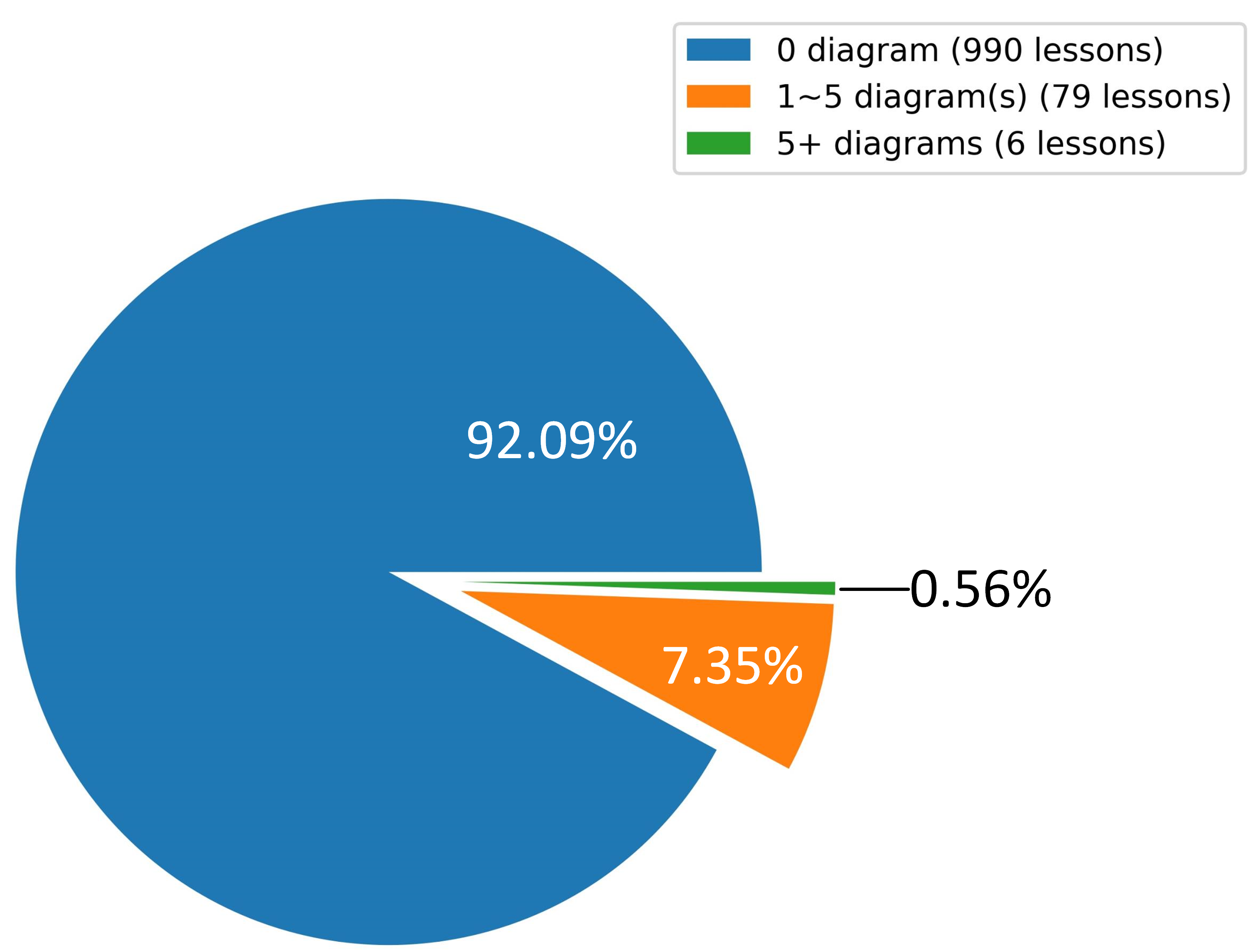}
	\caption{Statistics of instructional diagrams in all lessons. The diagram is the image included in multi-modal contexts of lessons.}
	\label{ins-dia}
\end{figure}
\begin{figure*}[tbp]
	\centering
	\includegraphics[width=\linewidth]{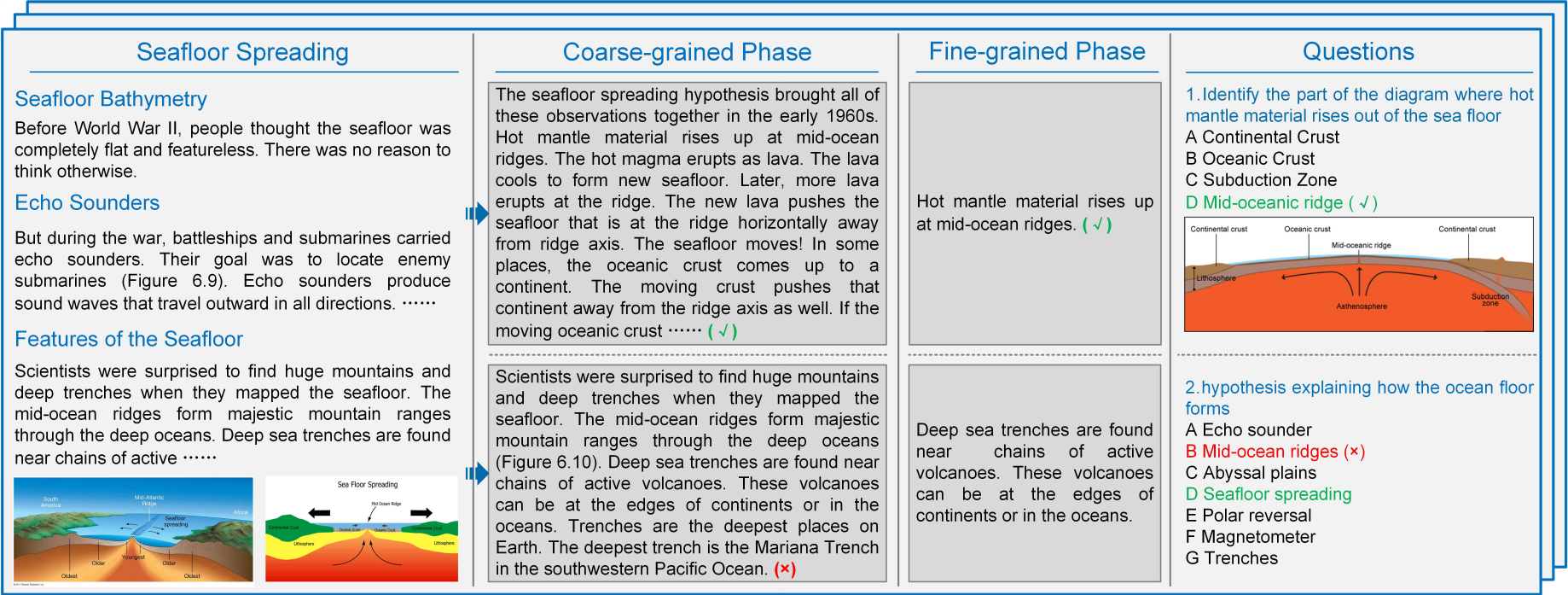}
	\caption{Case studies to show the strengths and weaknesses of XTQA. The left part shows the multi-modal context of lesson \emph{seafloor spreading}. The middle parts show the explanation extracting processes by our proposed algorithm. The right part shows the questions of this lesson. These cases come from the validation split.}
	\label{fig.3}
\end{figure*}

\subsection{Discussions}
We use the answer label to optimize the explanation extracting due to the lack of ground truth for explanations in CK12-QA, \emph{i.e.}, weakly supervised learning. To further analyze the effectiveness of this optimization, we conduct experiments on the first 150 non-diagram and diagram questions of the test split respectively. These questions have gold explanations and are introduced in Section \ref{explanation}. The results in Figure \ref{fig-span-acc} show the relationship between question answering and explanation extracting.

\noindent \textbf{Questions Answered Correctly} The explanation result (mIoU) on the non-diagram questions, diagram questions and total questions are $72.38\%$, $75.62\%$ and $73.65\%$ respectively. The result shows the evidence for most of the questions answered correctly can provide accurate explanations for humans and proves the validity of our optimization to some extent. It also shows that accurate span-level explanations play a key role in answering questions. 

\noindent \textbf{Questions Answered incorrectly} The explanation result (mIoU) on the non-diagram questions, diagram questions and total questions are $34.33\%$, $37.06\%$ and $35.97\%$ respectively. The result shows the other information such as diagram representation is also important to answer questions accurately, although there are accurate span-level explanations. 

We also conduct experiments to explain why our method does not use the instructional diagrams included in multi-modal contexts of lessons, although the diagram is important. First, only $\sim$4\% and $\sim$12.5\% of text and diagram questions need instructional diagrams respectively according to the statistics of \cite{kembhavi2017you}. Compared with this, $\sim$80\% of the questions need single ($\sim$40\%) or multiple sentences ($\sim$40\%). Secondly, the instructional diagram does not always exist in each lesson, which is not conducive to model training. The statistic result of instructional diagrams in Figure \ref{ins-dia} shows that 92.09\% of the lessons have no instructional diagrams. Considering the above situations, we think evidence span extraction is a trade-off way to provide explanations for students.

\subsection{Case Studies}
We conduct the case studies as shown in Figure \ref{fig.3} to present the strengths and weaknesses of our method intuitively.

\noindent \textbf{Strengths} XTQA can provide the explicit span-level explanations with different lengths for answering different questions. For example, XTQA provides the explanation of length $1$ of the diagram question, and provides the explanation of length $2$ of the non-diagram question for humans respectively as shown in the middle part of Figure \ref{fig.3}.

\noindent \textbf{Weaknesses} If the coarse grained algorithm makes errors, it will cause the failure to find span-level explanations in the fine-grained phase. For example, XTQA finds the wrong top $1$ paragraph for the non-diagram question as shown in the middle part of Figure \ref{fig.3}, which causes the failure to find the explanations.

\section{Conclusion}
\label{conclusion}
In this paper, we propose a novel architecture towards span-level eXplanations for Textbook Question Answering (XTQA). It takes into account the entire context of a lesson as candidate evidence and employs the proposed coarse-to-fine grained explanation extracting (EE) algorithm to extract span-level explanations with varying lengths for different questions. Experimental results show that XTQA obtains the best explanation result and achieves the state-of-the-art performance on the validation and test split of CK12-QA respectively. Experimental results also demonstrate that the EE algorithm can be integrated into other TQA methods to enable them to have explainability and improve their TQA performance.

In the future, the following directions will be explored.
\begin{itemize}
	\item Error reduction of the coarse-grained algorithm may improve the accuracy of explanation extracting. For example, we can fine-tune the large pre-trained language model to retrieve the closest paragraph. We will explore how to devise an end-to-end architecture to optimize the process of the coarse-grained explanation extracting. 
	
	\item External knowledge helps to improve the performance of other tasks such as named entity recognition and VQA. An analysis about CK12-QA \citep{kembhavi2017you} also shows that about $10\%$ of the questions require external knowledge to answer. We will investigate how to integrate the external knowledge into XTQA to improve its performance. 
\end{itemize}

\section*{Acknowledgments}
	This work was supported by National Key Research and Development Program of China (2022YFC3303603, 2021YFB1715600), National Natural Science Foundation of China (62137002, U22B2019, 62272372, 62293553, 62250066, and 621737002), Innovative Research Group of the National Natural Science Foundation of China (61721002). I am also deeply indebted to Junjun Li, Yi Huang, and Jianlong Zhou for their direct and indirect help to me.

\bibliographystyle{IEEEtran}
\bibliography{sample-base}

\begin{IEEEbiography}[{\includegraphics[width=1in,height=1.25in,clip,keepaspectratio]{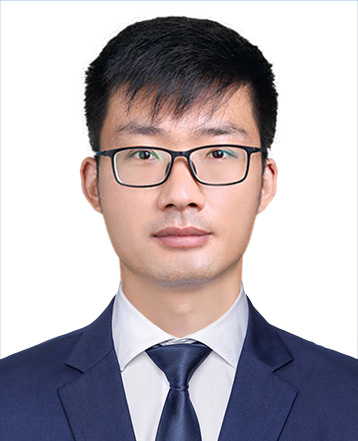}}]{Jie Ma}	
	(Member, IEEE) is an Assistant Professor in the School of Cyber Science and Engineering (Faculty of Electronic and Information Engineering) at Xi'an Jiaotong University, Xi'an, Shaanxi 710049, P.R. China. He is also a member of the Ministry of Education's Key Lab for Intelligent Networks and Network Security. His research interests cover natural language processing and trustworthy multimodality learning, focusing particularly on knowledge graph learning, robust visual question answering, and question dialogue. He has contributed to several top journals and conferences, including IEEE TIP, TNNLS, IJCAI, and WSDM. Furthermore, he has served as a program committee member for numerous conferences such as ICLR and AAAI, and reviewed manuscripts for multiple journals such as IEEE TIP and TNNLS. For more information, please visit \url{https://dr-majie.github.io/}.
\end{IEEEbiography}

\begin{IEEEbiography}[{\includegraphics[width=1in,height=1.25in,clip,keepaspectratio]{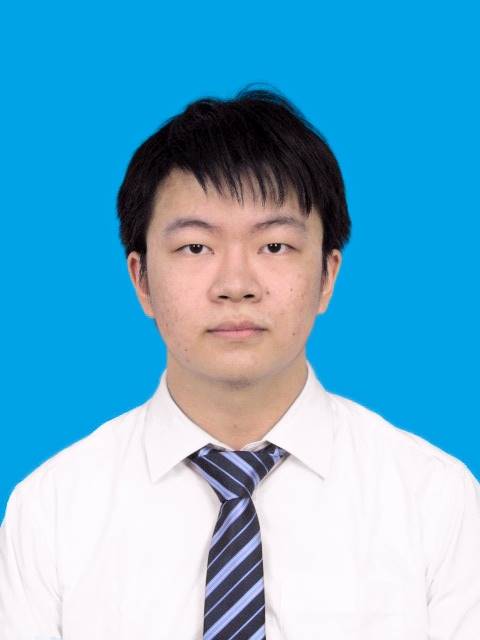}}]{Qi Chai}
	received his B.E. degree from Xi'an Jiaotong University, Xi'an, China, in 2021. He is currently pursuing the M.E. degree with the School of Computer Science and Technology at Xi'an Jiaotong University, Xi'an, China. His research interests include multi-modal question answering ,image processing and natural language processing.
\end{IEEEbiography}

\begin{IEEEbiography}[{\includegraphics[width=1in,height=1.25in,clip,keepaspectratio]{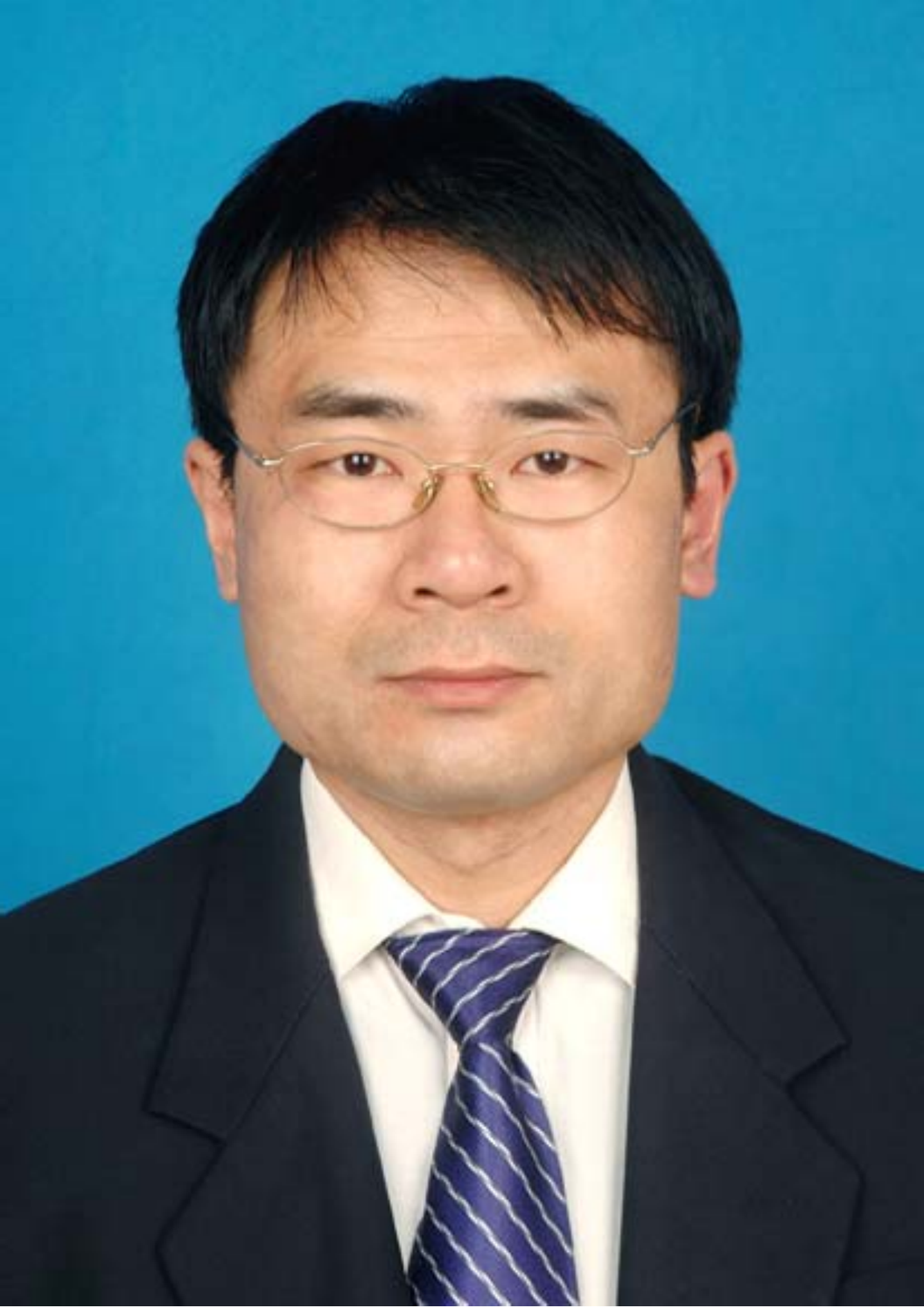}}]{Jun Liu}
	(Senior Member, IEEE) received the B.S. in computer science and technology in 1995 and Ph.D. degrees in systems engineering in 2004, both from Xi’an Jiaotong University, China. He is currently a Professor with the Department of Computer Science, Xi’an Jiaotong University. He has authored more than ninety research papers in various journals and conference proceedings. He has won the best paper awards in IEEE ISSRE 2016 and IEEE ICBK 2016. His research interests include NLP and e-learning. Dr. Liu currently serves as an associate editor of IEEE TNNLS from 2020, and has served as a guest editor for many technical journals, such as Information Fusion, IEEE SYSTEMS JOURNAL. He also acted as a conference/workshop/track chair at numerous conferences.
\end{IEEEbiography}

\begin{IEEEbiography}[{\includegraphics[width=1in,height=1.25in,clip,keepaspectratio]{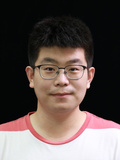}}]{Qingyu Yin}
	is a scientist in Amazon. He got his Ph.D. from Harbin Institute of Technology, advised by Prof. Ting Liu. His research interests lie in natural language processing and human-computer dialogue.
\end{IEEEbiography}

\begin{IEEEbiography}[{\includegraphics[width=1in,height=1.25in,clip,keepaspectratio]{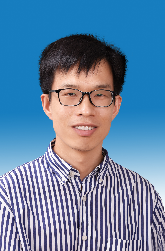}}]{Pinghui Wang}
	(Senior Member, IEEE) is currently a professor with the MOE Key Laboratory for Intelligent Networks and Network Security, Xi’an Jiaotong University, Xi’an, China, and also with the Shenzhen Research Institute, Xi’an Jiaotong University, Shenzhen, China. His research interests include internet traffic measurement and modeling, traffic classification, abnormal detection, and online social network measurement.
\end{IEEEbiography}

\begin{IEEEbiography}[{\includegraphics[width=1in,height=1.25in,clip,keepaspectratio]{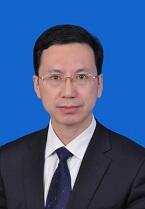}}]{Qinghua Zheng}
	 received the B.S. degree in computer software, the M.S. degree in computer organization and architecture, and the Ph.D. degree in system engineering from Xi’an Jiaotong University, Xi’an, China, in 1990, 1993, and 1997, respectively. He was a Postdoctoral Researcher with the Harvard University, Cambridge, MA, USA, in 2002. He is currently a Professor with the Department of Computer Science and Technology, Xi’an Jiaotong University. His research interests include knowledge engineering of big data and software trustworthiness evaluation.
\end{IEEEbiography}

\vfill

\end{document}